\documentclass[draft]{agujournal2019}
\usepackage{url} 
\usepackage{lineno}
\usepackage[inline]{trackchanges} 
\usepackage{soul}
\usepackage{amsmath}

\draftfalse

\journalname{Enter journal name here}

\begin{document}

%
%


\title{Eddy-Resolving Global Ocean Forecasting with Multi-Scale Graph Neural Networks}

%
%




\authors{Yuta Hirabayashi\affil{1}, Daisuke Matsuoka\affil{1}, Konobu Kimura\affil{2}}


\affiliation{1}{Japan Agency for Marine-Earth Science and Technology}
\affiliation{2}{FURUNO ELECTRONIC CO. LTD.}




\correspondingauthor{Yuta Hirabayashi}{yuta.hirabayashi@jamstec.go.jp}



\begin{keypoints}
\item A multi-scale graph neural network that uses two spherical meshes for 10-day eddy-resolving global ocean forecasts is proposed.
\item The model is designed to improve the representation of multi-scale ocean dynamics and short-term variability.
\item Evaluations show improved representation across a wide range of spatial scales and higher short-term skill.
\end{keypoints}

%
%

%
%


\begin{abstract}
      Research on data-driven ocean models has progressed rapidly in recent years; 
      however, the application of these models to global eddy-resolving ocean forecasting remains limited. 
      The accurate representation of ocean dynamics across a wide range of spatial scales remains a major challenge 
      in such applications. 
      This study proposes a multi-scale graph neural network-based ocean model 
      for 10-day global forecasting that improves short-term prediction skill 
      and enhances the representation of multi-scale ocean variability. 
      The model employs an encoder-processor-decoder architecture 
      and uses two spherical meshes with different resolutions to better capture the multi-scale nature of ocean dynamics.
      In addition, the model incorporates surface atmospheric variables along with ocean state variables 
      as node inputs to improve short-term prediction accuracy by representing atmospheric forcing.
      Evaluation using surface kinetic energy spectra and case studies shows 
      that the model accurately represents a broad range of spatial scales, 
      while root mean square error comparisons demonstrate improved skill in short-term predictions. 
      These results indicate that the proposed model delivers more accurate short-term forecasts 
      and improved representation of multi-scale ocean dynamics, 
      thereby highlighting its potential to advance data-driven, eddy-resolving global ocean forecasting.
\end{abstract}

\section*{Plain Language Summary}
      Ocean forecasting is essential for maritime operations, such as ship route planning and fisheries.
      Traditional numerical ocean models play a critical role; however, they require substantial computational resources. 
      To this end, data-driven approaches have demonstrated effectiveness across various fields.
      However, their application to high-resolution global ocean forecasting remains limited.
      This study develops a new data-driven method that learns from reanalysis data to predict ocean conditions.
      The data-driven method uses a machine learning model called a graph neural network, which is designed to process information on spherical grids covering the globe.
      By combining two different grid resolutions, the model can represent both large- and small-scale ocean features more consistently.
      In addition, the model incorporates information regarding winds and other surface atmospheric conditions, which improves short-term forecasts in surface layers.
      Results show that the proposed model improves short-term global ocean forecasting 
      and the representation of ocean variability across multiple spatial scales.
      These features are indispensable for high-resolution global ocean modeling, 
      highlighting the potential of data-driven global ocean forecasting models.

%
%

%


%
%
%
%
======

\section{Introduction}
High-resolution global ocean forecasting is essential for scientific research and practical applications, 
yet computationally challenging.
In particular, eddy-resolving ocean forecasting ($1/10^\circ$ or finer resolution) enables explicit representation of key oceanic features,  
including western boundary currents, mesoscale variability, and the position and sharpness of ocean fronts \cite{Hurlburt1996, Hurlburt1998, Metzger2014}.  
Furthermore, mesoscale variability influences plankton distribution \cite{McGillicuddy2016},  
which is closely tied to the location of fishing grounds \cite{Woodsona2015}. 
Therefore, eddy-resolving ocean state information contributes not only to advancing our understanding of ocean dynamics  
but also to supporting industries such as fisheries.
To provide accurate global ocean state information,  
ocean global circulation models (OGCMs) play a crucial role.
Although eddy-resolving OGCMs simulate ocean states more accurately than coarser-resolution models \cite{Hurlburt2008},
their increased spatial resolution results in substantially greater computational demands.

Data-driven approaches have gained prominence in computational fluid dynamics, including ocean forecasting, 
as they can achieve high accuracy while maintaining low computational costs.
These approaches utilize advanced machine learning models
trained to capture the evolution of fluid dynamics (e.g., \citeA{Roznowicz2024}).
In global weather forecasting, numerous data-driven models have been developed \cite{Keisler2022, Pathak2022, Nguyen2023, Chen2023, Bi2023, Lam2023, Lang2024},
several of which report outperforming the ECMWF Integrated Forecasting System (IFS), a state-of-the-art numerical global weather forecasting system.
Building on the success of these data-driven models in global weather forecasting,
similar approaches have been explored for global ocean forecasting.
For example, \citeA{Xiong2023} developed a data-driven ocean model based on the Fourier Neural Operator (FNO) \cite{Li2020} at $1/4^\circ$ resolution.
Subsequently, \citeA{ElAouni2025} proposed another FNO-based model also at $1/4^\circ$ resolution,
thereby introducing comprehensive validation metrics and demonstrating promising results.
These models primarily focus on short-term predictions (e.g., 10 days),
whereas global climate ocean emulators, such as \citeA{Guo2025, Dheeshjith2025, Subel2024},
have extended prediction timescales to the decadal range,
demonstrating the capability to stably emulate ocean dynamics over decadal periods
at $1^\circ$ resolution.

However, although data-driven ocean forecasting models successfully reproduce coarse-scale patterns at lower resolutions,
whether they can represent fine-scale structures remains unclear.
In fact, in weather forecasting, data-driven models have shown limitations in capturing fine-scale features compared with traditional numerical weather prediction models \cite{Bonavita2024}.
Given these considerations, the feasibility of data-driven models for eddy-resolving global ocean forecasting warrants investigation.
Motivated by this knowledge gap, several recent studies have proposed data-driven models for eddy-resolving global ocean forecasting \cite{Wang2024,Cui2025,Huang2025}.
For example, \citeA{Wang2024} proposed an architecture based on the Swin Transformer \cite{Liu2021} and
showed that their model outperforms several operational numerical ocean forecasting systems in terms of root mean square error (RMSE).
Subsequently, \citeA{Cui2025} demonstrated the effectiveness of incorporating atmospheric variables,
while \citeA{Huang2025} employed the mixture-of-time module to capture multi-scale temporal dependencies of ocean variables.
Although these studies demonstrate the potential of data-driven models for eddy-resolving global ocean forecasting,
the capability of these models to represent oceanic structures across a wide range of spatial scales is potentially constrained.
This is partly because these models rely on similar Swin Transformer-based architectures with rectilinear meshes that lack spherical symmetry, leading to unphysical inductive biases \cite{Linander2025, Ramavajjala2024}.
Such limitations may hinder accurate representation of the broad spatial scales required for eddy-resolving ocean models.

This study applies a multi-scale graph neural network (GNN) \cite{Fortunato2022}
to global ocean forecasting at $1/12^\circ$ resolution over a 10-day period.
The multi-scale GNN employs an encoder-processor-decoder architecture designed to capture the multi-scale spatial characteristics of ocean dynamics.
This architecture incorporates two spherical meshes with different spatial resolutions to naturally encode global information.
Thus, the proposed architecture is expected to represent a wide range of spatial scales, from mesoscale eddies to basin-scale features such as western boundary currents.
This multi-scale design is motivated by the success of GraphCast \cite{Lam2023} in global weather forecasting.
Building on GraphCast, this study tailors the architecture for eddy-resolving global ocean forecasting by removing land nodes and using only two mesh levels to focus on fine-scale dynamics.
In addition, to account for atmospheric forcing, surface atmospheric variables are combined with ocean variables and used as node input features, thereby improving the evolution of surface-layer fields that are particularly important for short-term prediction.
This study evaluates prediction accuracy and the ability of the model to capture ocean dynamics across a wide range of spatial scales, 
using quantitative and qualitative comparisons with the first data-driven, eddy-resolving global ocean forecasting model based on a Swin Transformer-based architecture proposed by \citeA{Wang2024}.

\section{Methods}
\subsection{Overview}   
\begin{figure}
      \centering
      \includegraphics[bb=0 0 718.08 585.12,clip,width=\linewidth]{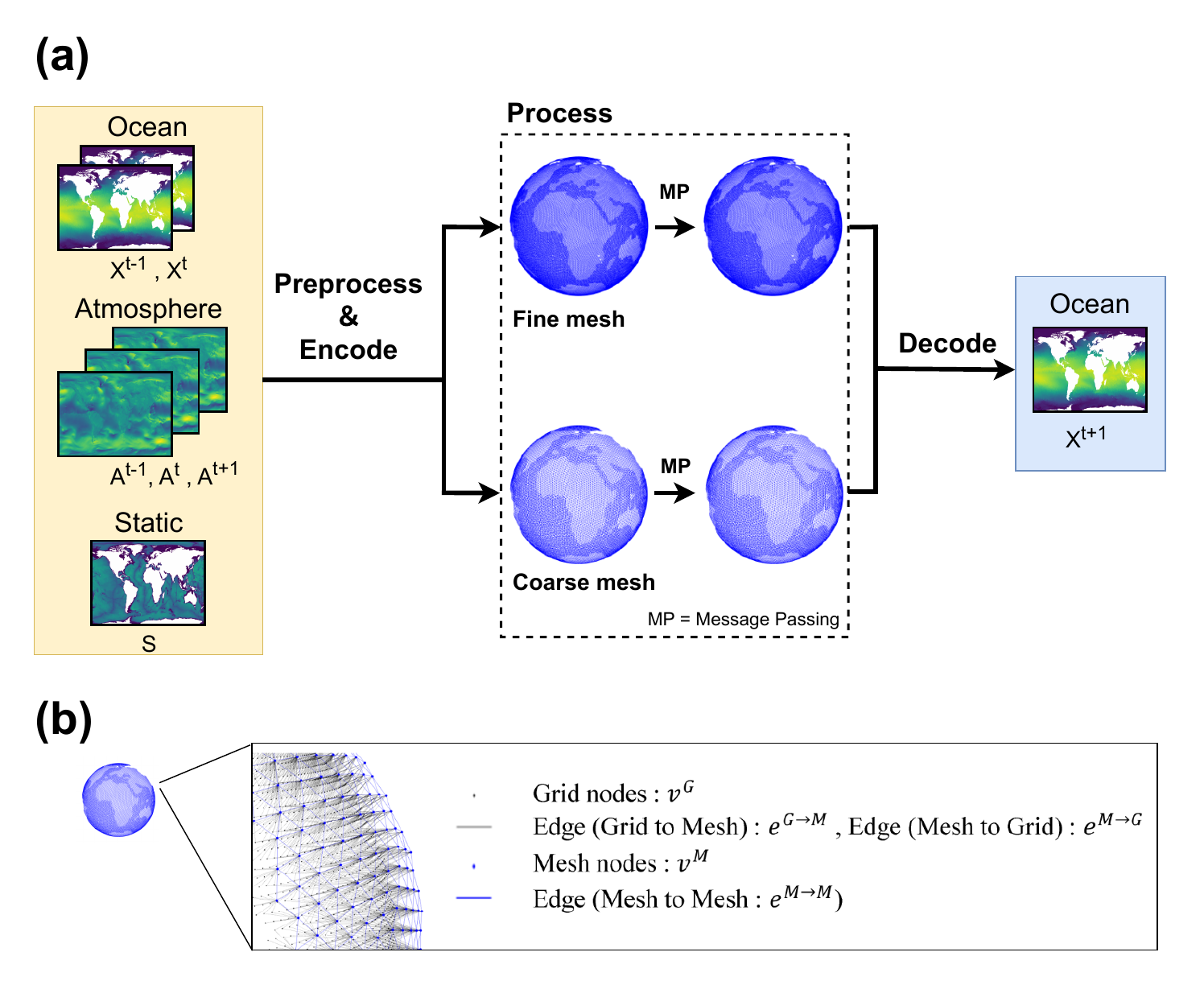}
      \caption{
            Overview of the proposed ocean-forecasting model.
            (a) Schematic of the model architecture, which takes ocean states ($X^{t-1}, X^t$), atmospheric conditions ($A^{t-1}, A^t, A^{t+1}$), and static features ($S$) as input. 
            These inputs are preprocessed and encoded, followed by independent message passing (MP) on coarse and fine meshes. 
            The encoded features are then decoded to predict the next ocean state ($X^{t+1}$).
            (b) Mesh structure, where grid nodes ($v^G$) and mesh nodes ($v^M$) are connected via directional edges: 
            grid-to-mesh ($e^{G\rightarrow M}$), mesh-to-grid ($e^{M\rightarrow G}$), and mesh-to-mesh ($e^{M\rightarrow M}$). 
            Grid-node positions align with those of the ocean reanalysis dataset (GLORYS).
      }
      \label{fig:Conceptual_Image}
\end{figure}

Figure \ref{fig:Conceptual_Image} (a) illustrates the overall schematic design of the proposed model.
The input consists of ocean state data at the current time step ($t = 0$) and one step prior ($t = -1$), 
denoted as $X^0$ and $X^{-1}$; 
atmospheric forecast data $A^{-1}, A^0, A^1$; 
and latitude, longitude, and ocean depth provided as static data $S$,
where $X_t$ represents the ocean state variables that include vertical levels,
whereas $A_t$ consists of atmospheric variables defined at the ocean surface.
Both $X_t$ and $A_t$ are treated as two-dimensional gridded data, with different depth levels in 
$X_t$ represented as separate variables, 
flattening the vertical dimension into the channel dimension of the input tensor.
Given these inputs, the multi-scale GNN model $F$ 
performs a one-step-ahead prediction of the ocean state,
as expressed in Equation~(\ref{eq:autoregressive}).
Forecasting is performed in an autoregressive manner up to $t = T$ 
by recursively feeding predicted ocean states back as input, 
with atmospheric data provided at each timestep.
In this study, we set $T = 10$, with each step corresponding to one day.
Accordingly, both ocean and atmospheric inputs are given as daily-mean fields.

\begin{equation}
      X^{t+1} = F(X^{t-1}, X^t, A^{t-1}, A^t, A^{t+1}, S)
      \label{eq:autoregressive}
\end{equation}

\subsection{Mesh construction}
The meshes used in this study are constructed according to the design described by \citeA{Lam2023}.
Among the hierarchical meshes defined in GraphCast, 
the two finest resolutions--the finest and second finest meshes--are employed to represent multi-scale ocean dynamics. 
As illustrated in Figure~\ref{fig:Conceptual_Image} (b), 
the graph includes grid nodes corresponding to the ocean data grid, mesh nodes at two types of resolutions, 
and edges connecting grid-to-mesh, mesh-to-grid, and mesh-to-mesh nodes. 

A key modification is introduced: land nodes are excluded from the graph structure. 
This refinement enables the mesh to focus exclusively on ocean regions, 
thereby improving both computational efficiency and physical relevance for ocean forecasting. 
This is implemented through a filtering process that removes land grid nodes and their associated edges from the graph. 
Specifically, edges connected to land nodes are excluded, 
and mesh-to-mesh connections are pruned to retain only those involving ocean-connected nodes, 
thus ensuring an efficient graph structure for ocean forecasting.

\subsection{Algorithm} \label{sec:algorithm}

This section outlines the prediction algorithm using a multi-scale GNN, 
which follows the overall methodology of GraphCast.

\noindent\textbf{Definition of GNN Operations}

The GNN operations in this study follow the message passing framework~\cite{Gilmer2017}, 
where the message function and update function are implemented using multilayer perceptrons (MLPs) 
with residual connections~\cite{Fortunato2022, Lam2023}:

\begin{equation}
    e_{s \rightarrow r} \leftarrow \mathrm{MLP}(e_{s \rightarrow r}, v_s, v_r) + e_{s \rightarrow r}
      \label{eq:edge_update}
\end{equation}

\begin{equation}
    v_r \leftarrow \mathrm{MLP} \left( v_r, \sum_{s \in \mathcal{N}(r)} e_{s \rightarrow r} \right) + v_r
      \label{eq:node_update}
\end{equation}

where $e_{s \rightarrow r}$ is the edge feature vector from node $s$ to node $r$, 
$v_r$ is the feature vector of node $r$, and $\mathcal{N}(r)$ denotes the set of neighbors of node $r$.
The operation $\text{GNN}_{N=n}(v_s, v_r, e_{s \rightarrow r})$ refers to applying 
equations (\ref{eq:edge_update}) and (\ref{eq:node_update}) for $n$ iterations.

\noindent\textbf{Preprocessing}

The features of the nodes $v^G$, $v^M$ and edges $e^{G \rightarrow M}$, $e^{M \rightarrow M}$, $e^{M \rightarrow G}$ are defined 
as shown in Fig.~\ref{fig:Conceptual_Image} (b). 
The node features $v^G$ are constructed by concatenating $X^{t-1}$, $X^t$, $A^{t-1}$, $A^t$, $A^{t+1}$, and $S$,
and $v^M$ represent positional vectors of the mesh nodes. 
The edge features are constructed using the relative positions of the edge endpoints, following the method outlined by \citeA{Lam2023}. 
Subsequently, all features are normalized.

\noindent\textbf{Encode}

To unify the embedding dimension, 
the features $v^G$, $v^M$, $e^{G \rightarrow M}$, $e^{M \rightarrow M}$, and $e^{M \rightarrow G}$ are individually 
processed by MLPs and projected into a common feature space. The embedding dimension is set to 192.

\begin{equation}
    v^M \leftarrow \mathrm{GNN}_{N=1}(v^M, v^G, e^{G \rightarrow M})
\end{equation}

\begin{equation} 
    v^G \leftarrow \mathrm{MLP}(v^G) + v^G
\end{equation}

\noindent\textbf{Process}

Using the updated $v^M$ and $e^{M \rightarrow M}$, 
GNN operations are applied to each mesh, where
the iteration number of message passing is set to 16.

\begin{equation}
    v^M \leftarrow \mathrm{GNN}_{N=16}(v^M, v^M, e^{M \rightarrow M})
\end{equation}

\noindent\textbf{Decode}

Based on $v^M$ and $e^{M \rightarrow G}$, the final grid-level features are computed as follows:

\begin{equation}
    v^G \leftarrow \mathrm{GNN}_{N=1}(v^G, v^M, e^{M \rightarrow G})
\end{equation}

Finally, the output is obtained by passing $v^G$ through an MLP and adding the result to $X^t$, thus producing the forecast $X^{t+1}$.

\subsection{Data and training settings} \label{sec:data_train}
Gridded ocean and atmospheric datasets were used for both training and evaluation.
For the ocean data, GLORYS12V1 \cite{Jean-Michel2021} (hereafter, referred to as GLORYS), a global ocean reanalysis dataset,
was used for $X^t$ and $S$,
with 23 vertical levels ranging from 0.49~m to 643.57~m.
For the atmospheric data at the ocean surface,
ECMWF’s ERA5 \cite{Hersbach2020}, a global atmospheric reanalysis dataset, was used for $A^t$ during training,
while NOAA’s GFS (Global Forecast System) forecast data, an operational atmospheric forecasting dataset, was used for evaluation.
Both ocean and atmospheric variables were provided as daily-mean fields,
and the atmospheric data were interpolated onto the ocean grid using bicubic interpolation.
A summary of the datasets used in this study is presented in Table~\ref{tab:variables}.

The model was trained using 25 years of reanalysis datasets in two training phases.
The training period spanned from January 1, 1993, to December 31, 2017.
In the first phase, the model was trained with 1-day predictions,
followed by fine-tuning with 2-day rollouts in the second phase.
Both training phases employed the AdamW optimizer \cite{Loshchilov2019} with $\beta_1 = 0.9$ and $\beta_2 = 0.95$, 
which are the coefficients used to compute the running averages of the gradient and its square, respectively,
and used the mean squared error (MSE) loss.
The learning rates were set to $10^{-3}$ and $10^{-4}$ for the first and second phases, respectively.
All training was conducted using eight NVIDIA A100 GPUs (40~GB memory each).

\section{Experiments}
\subsection{Experimental setting} \label{sec:evaluations}
Quantitative and qualitative evaluations were conducted in comparison with \citeA{Wang2024}, which serves as the baseline.
The RMSE evaluation was performed using 30 cases in 2019, 
with predictions initialized every 12 days starting on January 1, 2019,
using GLORYS as common initial conditions.
This RMSE comparison was conducted to assess the overall performance of both models;
however, RMSE alone does not capture spatial or temporal characteristics.
To evaluate these characteristics, surface kinetic energy spectra 
and case study analyses were conducted using forecasts initialized on January 1, 2019.
The surface kinetic energy spectra quantitatively assess the representation of fine-scale structures,
while the case study analyses qualitatively reveal both spatial and temporal characteristics.

In addition to comparison with the baseline, 
sensitivity experiments were conducted to investigate whether the proposed model appropriately reflects atmospheric forcing
by using different types of atmospheric input data. 
Following previous sensitivity studies in numerical ocean models \cite{Hurlburt2008, Metzger2014}, 
the input atmospheric data during the prediction phase were varied from forecasts (GFS) to reanalysis data (ERA5) and climatology. 
The climatological data were obtained as the daily mean values of ERA5 over the training period. 
In all cases, the initial ocean condition was fixed and obtained from GLORYS.
Evaluations were performed on the same 30 cases in 2019, with predictions initiated every 12 days.
As the accuracy of atmospheric data is expected to be highest for reanalysis, followed by forecast and then climatology,
the prediction accuracy of the developed model was expected to follow this order if the atmospheric inputs were properly utilized.

\subsection{Baseline} \label{sec:baseline}
The Swin Transformer-based model proposed by \citeA{Wang2024} was used as a baseline.
The baseline model can be directly compared with the model developed here, 
as its trained weights are publicly available (see Data Availability Statement), 
and several experimental settings are shared between the two studies.
Specifically, the baseline model was trained using the same ocean dataset (GLORYS) 
over the same period (1993--2017) and with the same loss function (MSE) as adopted in this study.
In addition, both models predict the same set of ocean variables at the same depth levels 
with a horizontal resolution of $1/12^\circ$.

Despite several shared settings, 
methodological differences other than model architecture exist 
between the baseline and this study.
The baseline trains separate models for each lead time, 
in contrast with the autoregressive approach adopted in this study.
In addition, atmospheric forcing is treated differently. 
The baseline model relies solely on the atmospheric wind field at the initial time, 
whereas the model presented in this study employs time-evolving atmospheric forecasts 
to represent surface boundary conditions, consistent with standard practices in numerical ocean modeling.
Experiments were conducted to investigate the influence of these methodological differences
and differences in model architecture on the predictive characteristics of the proposed model.

\section{Results}
\subsection{Spatially averaged prediction skill across lead times and depths}
The proposed model exhibited lower RMSE at the surface layer for short lead times (Figure~\ref{fig:GFS_VS_XIHE_ALL_Surface}).
At the first forecast step (1-day lead time), the RMSE of the proposed model was lower than that of the baseline.
Although RMSE increased with lead time for both models, the increase was more rapid for the proposed model than for the baseline.
Consequently, a relatively low RMSE was maintained for the proposed model up to a 4-day lead time.
Similar behavior was observed for all surface variables, including salinity, temperature, eastward current, northward current, and sea surface height.

\begin{figure}
      \centering
      \includegraphics[bb=0 0  1880.4 1988.4,clip,width=\linewidth]{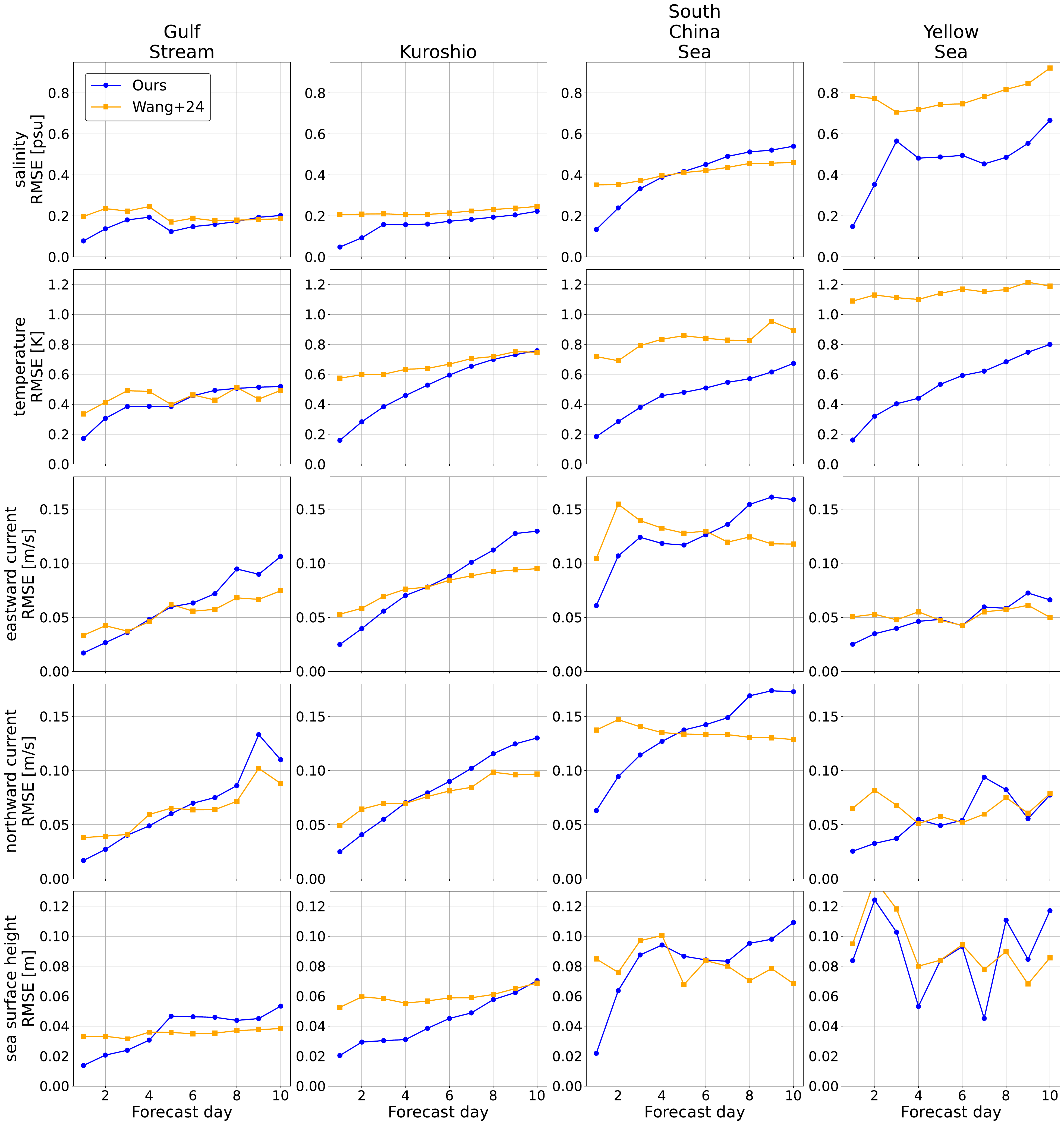}
      \caption{
            Comparison of RMSE over a 10-day forecast horizon for five surface variables, 
            salinity, temperature, eastward current, northward current, and sea surface height,
            computed against GLORYS.
            The evaluation is performed across four regional domains:
            Gulf Stream (76°W--40°W, 35°N--45°N), Kuroshio Extension (120°E--179°E, 20°N--55°N), 
            South China Sea (100°E--122°E, 0°N--27°N), and Yellow Sea (118°E--127°E, 30°N--42°N).
            The performance of the proposed model (Ours, blue) is compared with that of \citeA{Wang2024} (Wang+24, orange).
            The domain separation follows \citeA{Metzger2014}.
      }
      \label{fig:GFS_VS_XIHE_ALL_REGION_Surface}
\end{figure}

This short-lead-time surface skill was consistently observed across multiple ocean basins.
Figure~\ref{fig:GFS_VS_XIHE_ALL_REGION_Surface} shows comparisons over several regions, 
following the domain definitions of \citeA{Metzger2014}.
Similar trends to those in Figure~\ref{fig:GFS_VS_XIHE_ALL_Surface} were identified for the Kuroshio Extension, 
Gulf Stream, and South China Sea regions.
In contrast, in the Yellow Sea, where strong short-term variability is present \cite{Metzger2014}, 
both models exhibited relatively poor accuracy in sea surface height from the first forecast step.
Despite this degradation, lower RMSE at short lead times was consistently observed for the proposed model across regions.

\begin{figure}
      \centering
      \includegraphics[bb=0 0 1111.76 1448.4,clip,width=\linewidth]{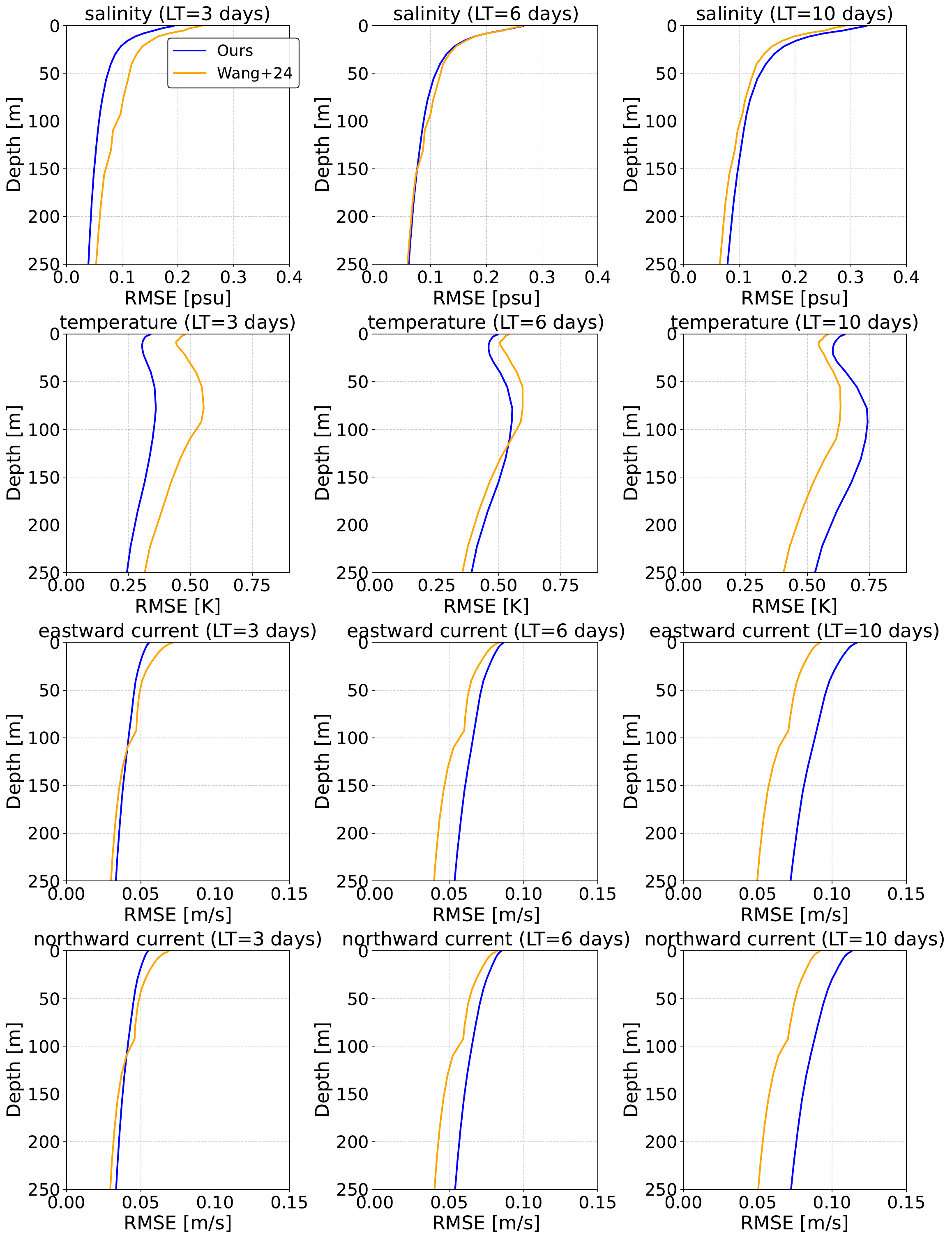}
      \caption{
            Comparison of RMSE profiles
            across depth for four ocean variables at forecast lead times of 3, 6, and 10 days computed against GLORYS.
            Each row corresponds to a different variable, and each column represents a different lead time.
      }
      \label{fig:VS_XIHE_Forcing_DEPTH_DAY}
\end{figure}

This short-lead-time improvement was also observed below the surface, particularly within the upper ocean layers.
Figure~\ref{fig:VS_XIHE_Forcing_DEPTH_DAY} shows that, for the eastward and northward currents at a 3-day lead time, 
the proposed model exhibited lower RMSE in the upper layers, 
approximately down to 100~m, whereas the baseline showed lower RMSE at greater depths.
At the 6- and 10-day lead times, 
RMSE of the proposed model increased across all depth layers.
For temperature and salinity, lower RMSE was observed for the proposed model across all depth layers at the 3-day lead time.
At the 6-day lead time, lower RMSE was mainly retained in the upper layers, 
while at the 10-day lead time, higher RMSE was observed across all layers.
Although the lead time at which RMSE degradation became evident varied among variables, 
lower RMSE at short lead times was observed on the surface and within the upper mixing layer.

\subsection{Evaluation of representation across spatial scales} \label{sec:spectra}
The surface kinetic energy (KE) spectra of the proposed model showed closer alignment with the reanalysis data (Figure~\ref{fig:VS_XiHe_FFT}). 
Specifically, for forecasts with a 1-day lead time, 
the KE spectra of the proposed model more closely matched those of GLORYS, 
whereas the baseline exhibited a deficiency at high wavenumbers in both the North Pacific and the North Atlantic.
At a 10-day lead time, a discrepancy emerged at high wavenumbers; 
however, the KE spectra of the proposed model remained closer to those of GLORYS than those of the baseline.
Overall, the spectral agreement was observed for both ocean basins at both lead times.

\subsection{Qualitative evaluation of spatiotemporal variability}
The proposed model produced surface temperature fields with finer-scale structures compared with the baseline (Figure~\ref{fig:temperature_2019-01-11}).
These comparisons were conducted over three western boundary current regions: 
the Agulhas Current, the Kuroshio Current, and the Gulf Stream.
These regions contain abundant mesoscale eddies and 
are therefore suitable for qualitative evaluation of the representation of fine-scale structures.
Across all regions, predictions from the proposed model exhibited discrepancies relative to GLORYS, 
reflecting the inherent difficulty of forecasting in these regions.
Nevertheless, the proposed model produced temperature fields that preserved fine-scale structures 
more clearly than those of the baseline.
This behavior is consistent with the spectra-based evaluation shown in Figure~\ref{fig:VS_XiHe_FFT}.

Beyond spatial structures, 
differences in the temporal evolution of fine-scale features were observed 
between the proposed model and the baseline (Figure~\ref{fig:VS_XiHe_India}).
By comparing sea surface height and surface currents in the Indian Ocean at 1-, 5-, and 10-day lead times, 
the temporal evolution of eddy structures was identified in GLORYS.
At a 5-day lead time, the proposed model reproduced the temporal development of a coherent eddy structure (red boxes), 
whereas the baseline did not reproduce this feature.
At a 10-day lead time, erroneous developments were observed in the proposed model (orange boxes),
which are further examined in the sensitivity experiments.

\subsection{Sensitivity analysis to input atmospheric forcing} \label{sec:sensitivity}
\begin{figure}
      \centering
      \includegraphics[bb=0 0 1111.76 1448.4,clip,width=\linewidth]{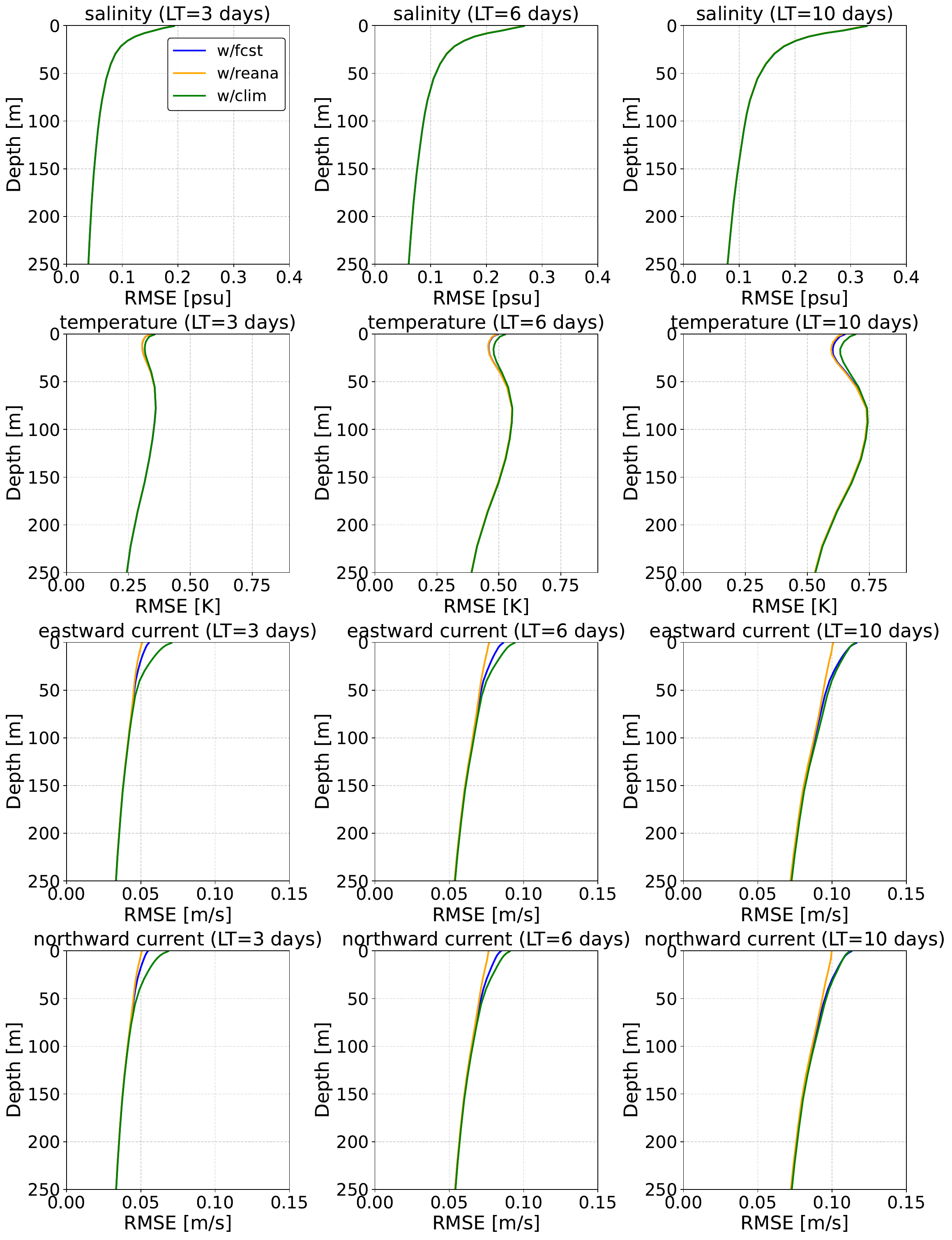}
      \caption{
            Sensitivity analysis of prediction accuracy across depth under different atmospheric forcing:
            Each row corresponds to a different variable, 
            and each column corresponds to a different forecast lead time (3, 6, and 10 days). 
            The horizontal axis shows the RMSE computed against GLORYS, 
            and the vertical axis indicates the depth range evaluated. 
            The blue curves (w/fcst) represent predictions forced by GFS forecast data, 
            the orange curves (w/reana) represent predictions forced by ERA5 reanalysis data, 
            and the green curves (w/clim) represent predictions forced by ERA5 climatological data.
      }
      \label{fig:Forcing_DEPTH_DAY}
\end{figure}

The influence of atmospheric input on the predictions was examined through sensitivity experiments.
In these experiments, atmospheric forecast data were replaced with reanalysis and climatological data, 
following similar sensitivity studies in numerical ocean models \cite{Hurlburt2008,Metzger2014}.
Figure~\ref{fig:Forcing_surf} shows that atmospheric forcing had a strong impact on temperature and current predictions 
at short lead times, whereas its impact on salinity and sea surface height was relatively limited.
The RMSE associated with atmospheric forecast forcing increased rapidly with lead time 
and eventually approached the RMSE obtained using climatological forcing.
In addition, the influence of atmospheric forcing was most pronounced in the upper ocean layers (Figure~\ref{fig:Forcing_DEPTH_DAY}).
Overall, the influence of atmospheric input was evident for temperature and current fields at short lead times and near the surface.

\begin{figure}
      \centering
      \includegraphics[bb=0 0 863.564 446.836,clip,width=\linewidth]{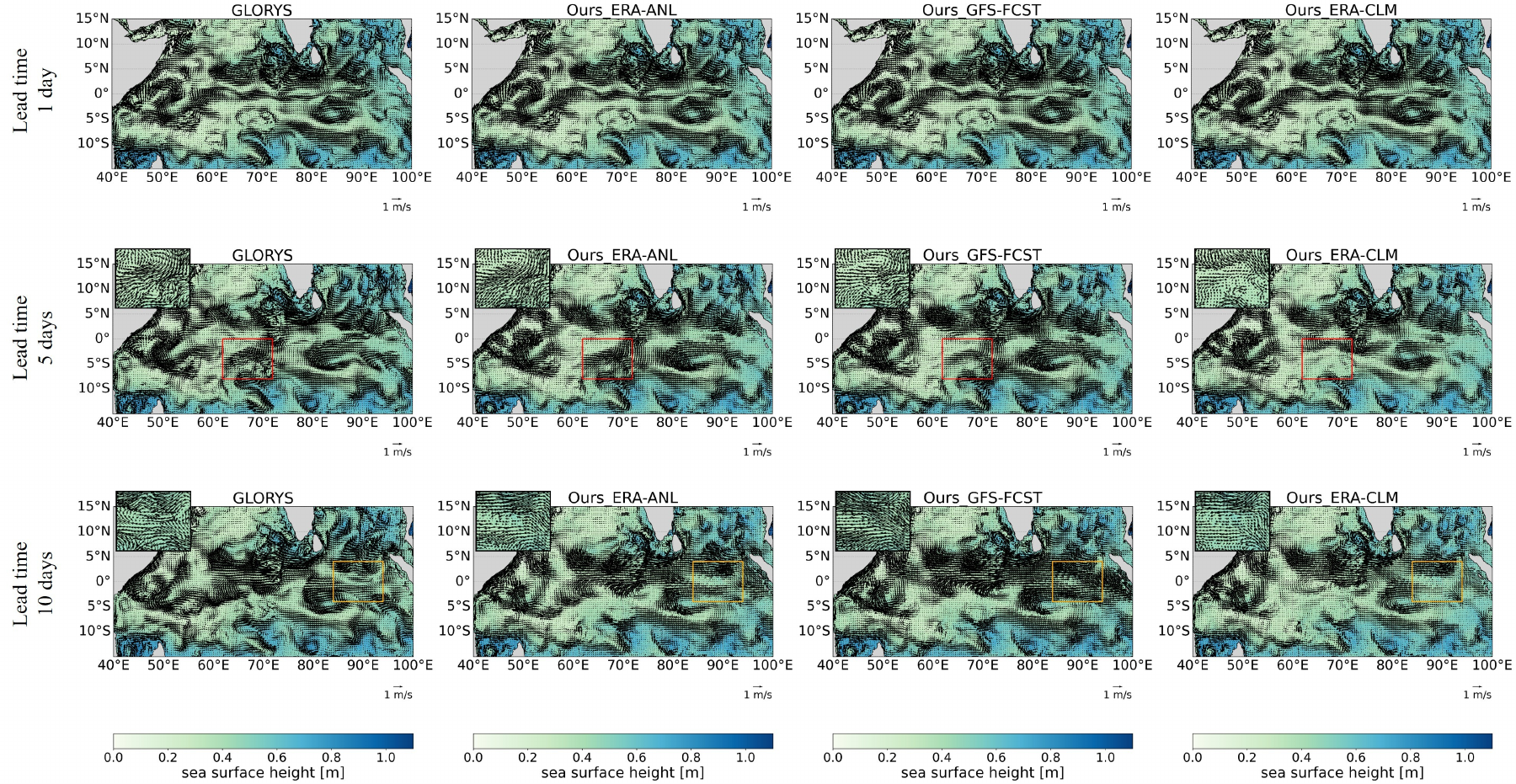}
      \caption{
            Comparison of sea surface height and surface currents in the Indian Ocean using different atmospheric forcing:
            Sea surface height and surface currents (vector fields) are shown for the Indian Ocean, 
            based on the reanalysis dataset (GLORYS, leftmost column) and the proposed model with three different atmospheric forcings: 
            ERA5 reanalysis (ERA-ANL), GFS forecast (GFS-FCST), and ERA5 climatology (ERA-CLM). 
            From top to bottom, the rows correspond to forecast days 1, 5, and 10, respectively.
            Orange dashed boxes highlight regions where the prediction accuracy differs depending on the atmospheric forcing.
            Inset panels provide magnified views of these regions.
      }
      \label{fig:FORCING_India}
\end{figure}

The effects of atmospheric forcing were further illustrated qualitatively through a case study (Figure~\ref{fig:FORCING_India}).
This case study was conducted for the region corresponding to Figure~\ref{fig:VS_XiHe_India}.
At a 10-day lead time, degradation was evident when atmospheric forecast forcing was used (orange boxes);
however, this degradation was reduced when atmospheric reanalysis forcing was applied.
At a 5-day lead time, the temporal development of the surface current field was reproduced
when atmospheric forecast forcing was applied (as shown in Figure~\ref{fig:VS_XiHe_India}), 
while degraded structures were observed when atmospheric climatological forcing was used (red boxes).
Overall, differences in the temporal evolution of surface currents were consistently observed depending on the atmospheric forcing used.

\section{Discussion and Conclusions}
\begin{figure}
      \centering
      \includegraphics[bb=0 0 1448.4 1304.4,clip,width=\linewidth]{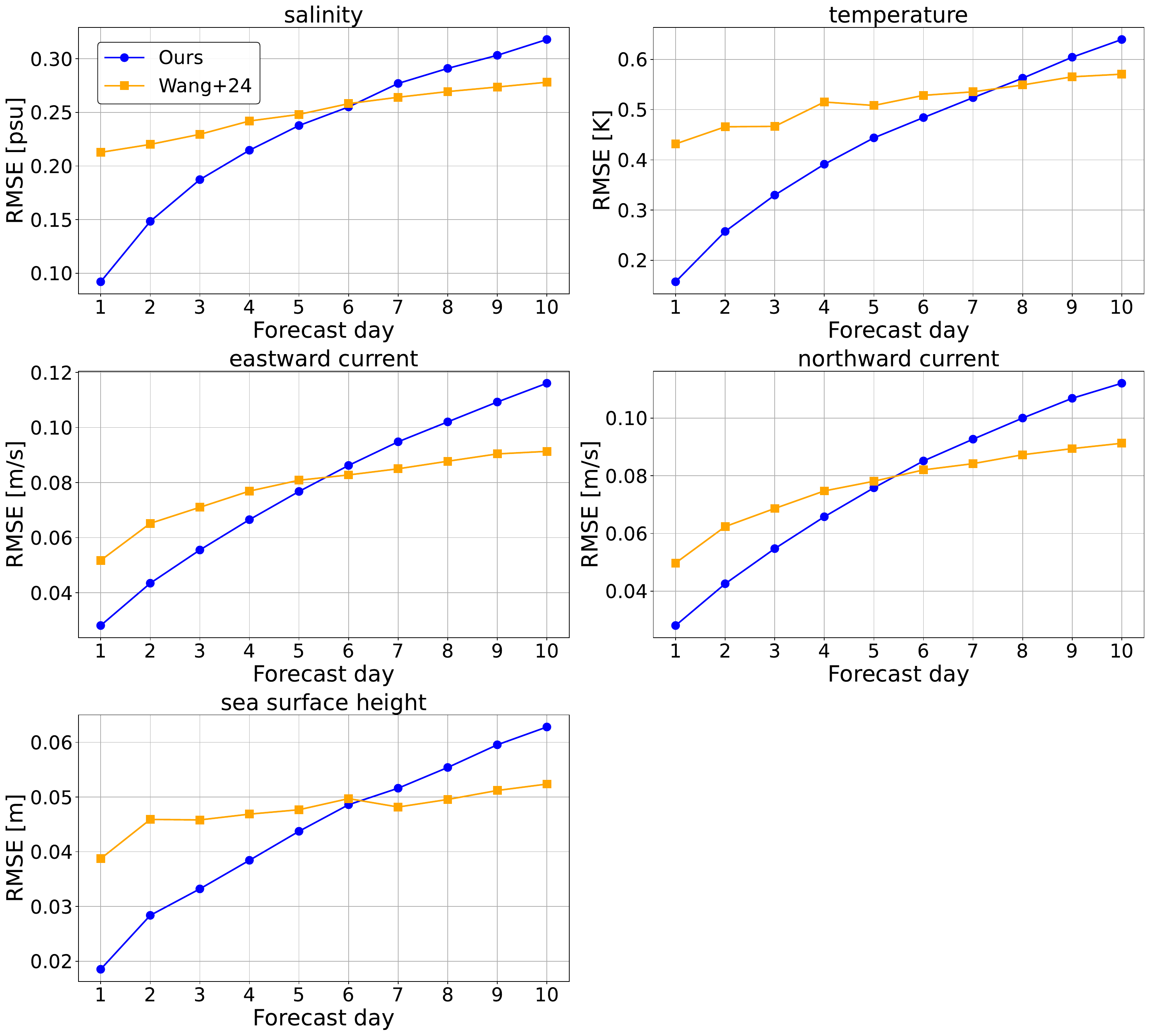}
      \caption{
            Comparison of RMSE over a 10-day forecast horizon for five surface variables, 
            computed against GLORYS.

      }
      \label{fig:GFS_VS_XIHE_ALL_Surface}
\end{figure}

\begin{figure}
      \centering
      \includegraphics[bb=0 0 716.727 487.855,clip,width=\linewidth]{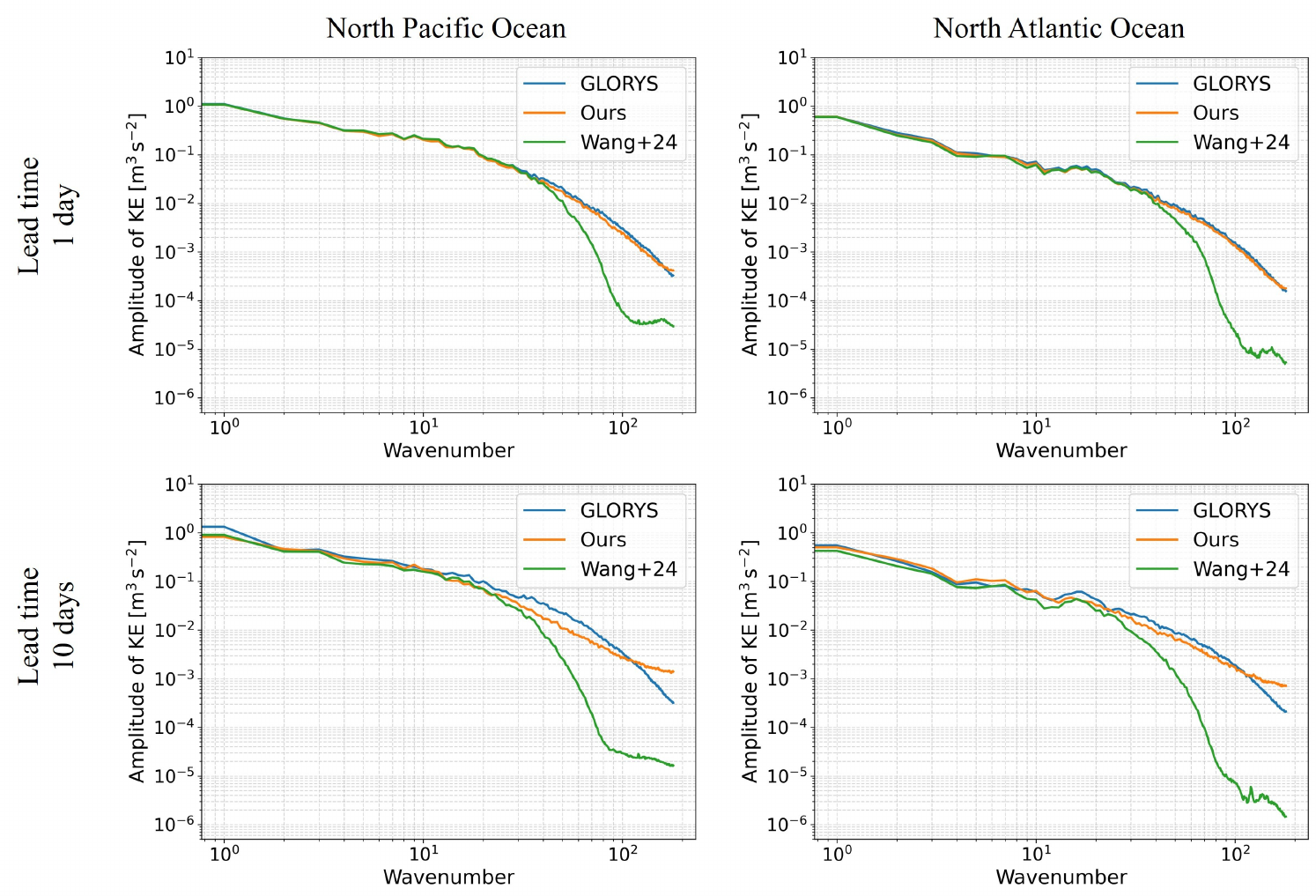}
      \caption{
            Kinetic energy spectra of surface velocity fields, presented as the amplitude of KE versus wavenumber.
            The left column shows results for the North Pacific region 
            \((10^\circ\text{--}40^\circ\text{N},~145^\circ\text{--}175^\circ\text{E})\), 
            while the right column shows results for the North Atlantic region 
            \((10^\circ\text{--}40^\circ\text{N},~60^\circ\text{--}30^\circ\text{W})\).
      }
      \label{fig:VS_XiHe_FFT}
\end{figure}

\begin{figure}
      \centering
      \includegraphics[bb=0 0  2111.76 1240.08,clip,width=\linewidth]{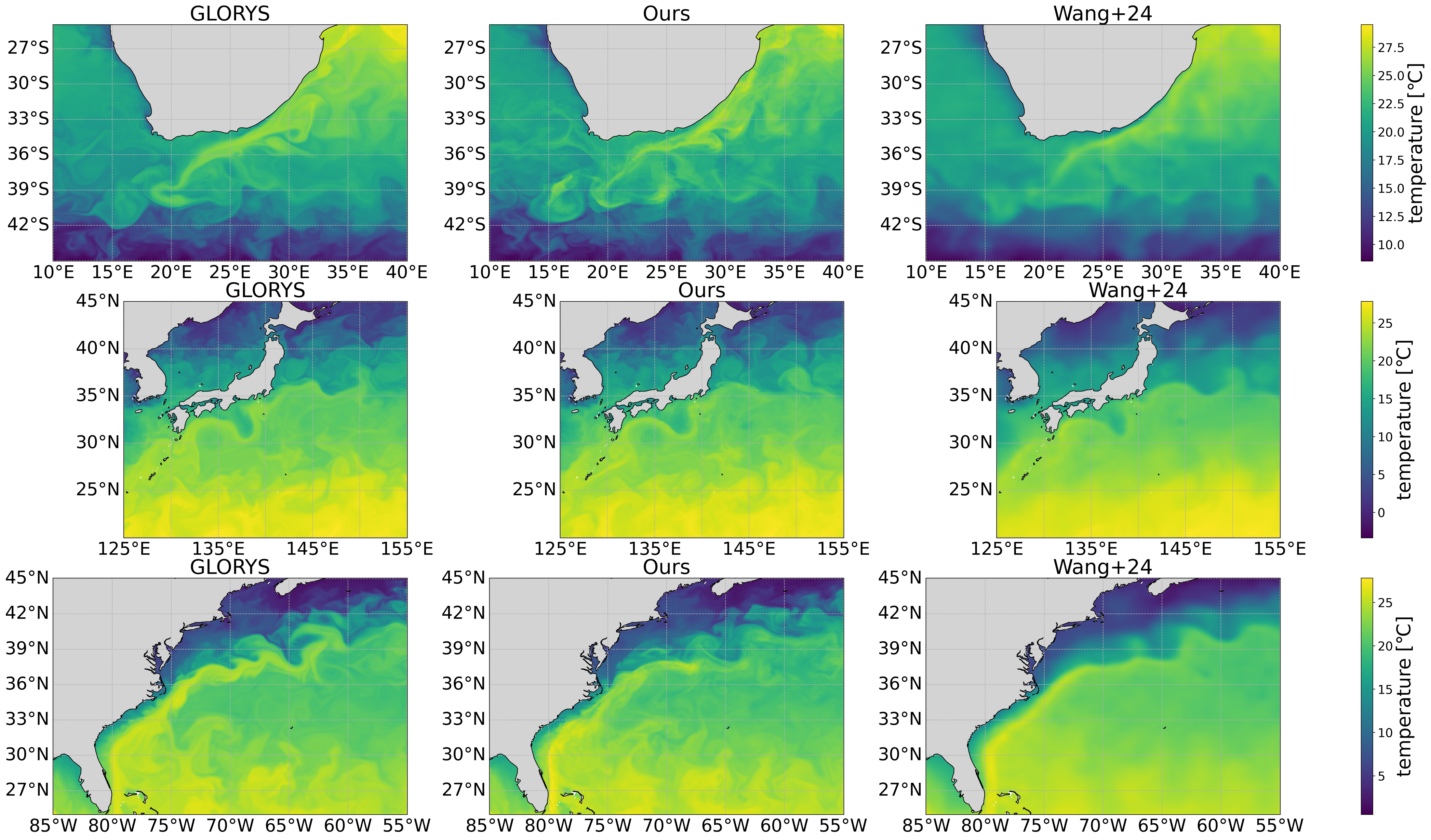}
      \caption{
            Comparison of sea surface temperature predictions at forecast day 10 in three regions:
            Sea surface temperature is shown for the Agulhas Current (top row), 
            Kuroshio Current (middle row), and the Gulf Stream (bottom row). 
            The results are presented for the reanalysis dataset (GLORYS, left), 
            the proposed model (Ours, center), and \citeA{Wang2024} (right).
      }
      \label{fig:temperature_2019-01-11}
\end{figure}

\begin{figure}
      \centering
      \includegraphics[bb=0 0 862.909 508.145 ,clip,width=\linewidth]{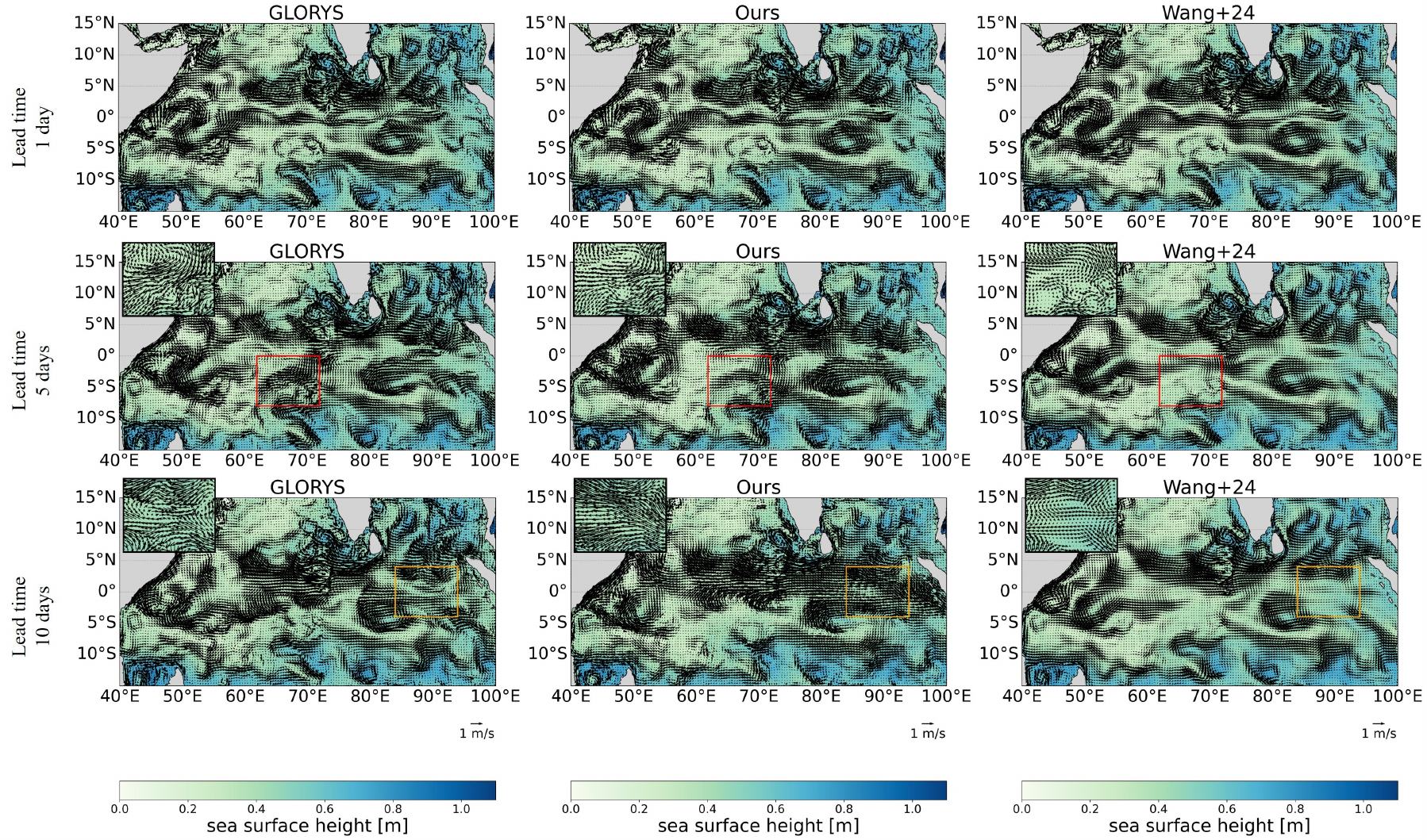}
      \caption{
            Comparison of sea surface height and surface currents in the Indian Ocean at different forecast lead times:
            Sea surface height and surface currents (vector fields) are shown for the Indian Ocean region 
            using the reanalysis dataset (GLORYS, left), the proposed model (Ours, center), 
            and \citeA{Wang2024} (right). 
            From top to bottom, the rows correspond to forecast days 1, 5, and 10, respectively.
            Red (orange) boxes highlight areas where the proposed model successfully (unsuccessfully) reproduced features.
            Inset panels provide magnified views of these highlighted regions.
      }
      \label{fig:VS_XiHe_India}
\end{figure}

\begin{figure}
      \centering
      \includegraphics[bb=0 0  1448.4 1304.4, clip,width=\linewidth]{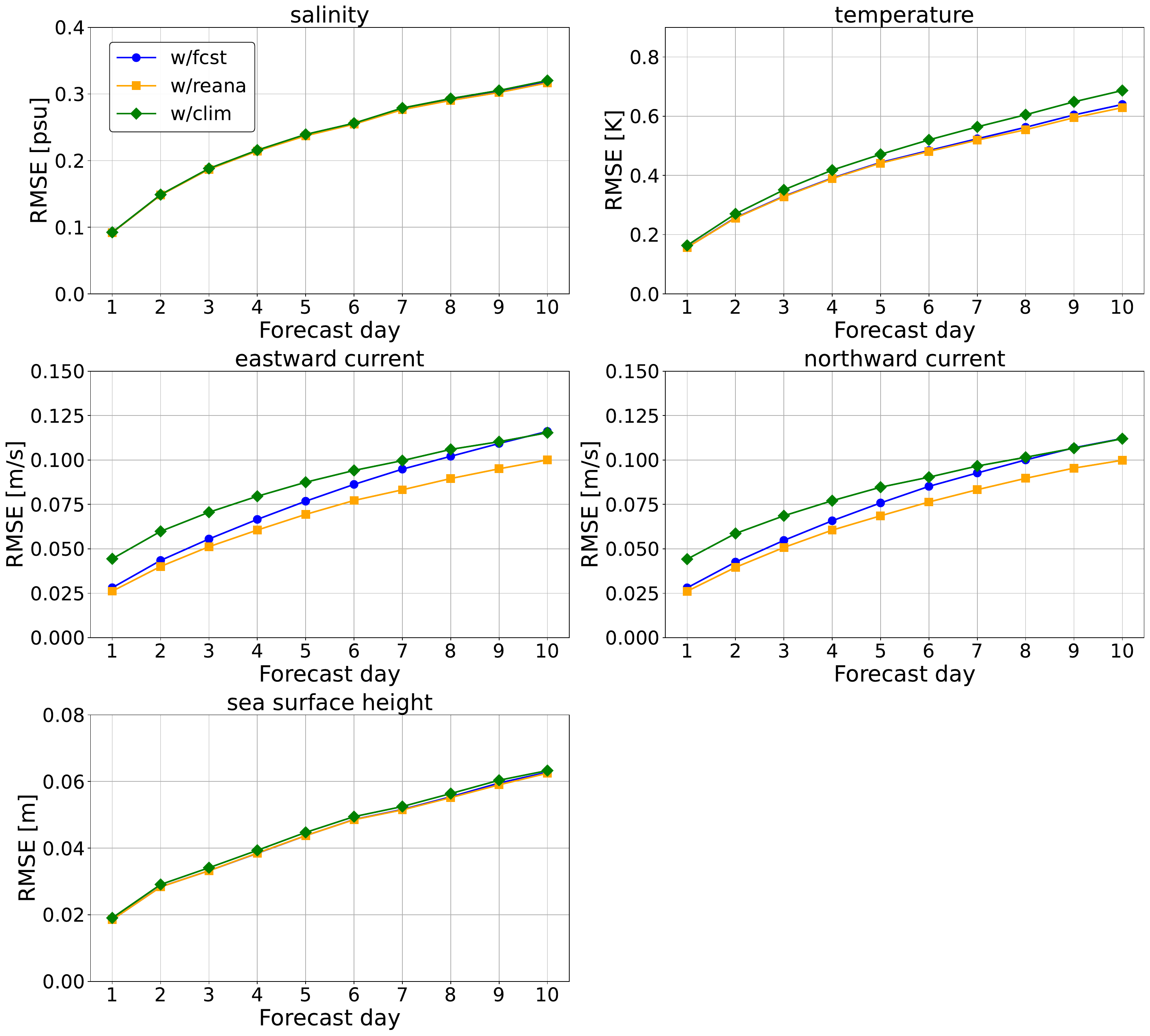}
      \caption{
            Sensitivity analysis of surface prediction accuracy to different atmospheric forcing:
            The horizontal axis represents the forecast lead time, and the vertical axis shows the RMSE 
            computed against GLORYS . 
            Each panel corresponds to a different variable 
            (salinity, temperature, eastward current, northward current, and sea surface height). 
            The blue circles (w/fcst) denote predictions forced by GFS forecast data, 
            the orange squares (w/reana) denote predictions forced by ERA5 reanalysis data, 
            and the green diamonds (w/clim) denote predictions forced by ERA5 climatological data.
      }
      \label{fig:Forcing_surf}
\end{figure}

This study aimed to develop a multi-scale GNN-based ocean prediction model that represents a wide range of spatial scales 
and achieves improved short-term skill, particularly within the upper ocean layers.
This representation capability is demonstrated by
kinetic energy (KE) spectra (Figure~\ref{fig:VS_XiHe_FFT}) and qualitative evaluations (Figure~\ref{fig:temperature_2019-01-11}, Figure~\ref{fig:VS_XiHe_India}),
demonstrating that the proposed model enables the reproduction of high spatial frequency components.
Specifically, the advantage of the proposed model in the KE spectra at the one-day forecast
is likely attributable to its architectural design,
as the spectra are not affected at the first prediction step by the choice of time evolution scheme.
Furthermore, the superiority in short-term and surface predictions (Figure~\ref{fig:GFS_VS_XIHE_ALL_Surface}) 
can be attributed to the incorporation of atmospheric forcing.
This interpretation is supported by the sensitivity experiments (Figure~\ref{fig:Forcing_surf}), 
which demonstrated that replacing atmospheric forecast forcing with reanalysis or climatological forcing led to systematic changes in prediction accuracy, particularly for temperature and surface currents at short lead times.
These findings indicate that the proposed model effectively utilizes atmospheric forcing to constrain short-term ocean evolution.
This behavior is consistent with established results from numerical ocean models, which emphasize the critical role of accurate atmospheric forcing in short-term ocean forecasting \cite{Hurlburt2008}.

The degradation of KE spectra representation at a lead time of 10 days indicates that 
extending the lead time further could make the proposed model outputs unstable. 
This instability is likely attributed to the proposed model’s autoregressive approach, where errors accumulate during rollout.
To address such issues, some studies have proposed incorporating additional loss terms to prevent instability, 
such as kinetic energy-based loss contributions \cite{Subel2024}. 
Although these treatments are effective for relatively low-resolution models, 
the feasibility of achieving stable long-lead forecasts at eddy-resolving resolution remains uncertain. 
The application of such approaches is beyond the scope of the present study, 
which focuses on 10-day forecasting; however, future research should consider 
applying them to the proposed model.

The RMSE degradation of the proposed model at longer lead times can be attributed to two factors: 
the sensitivity of RMSE to positional errors and error accumulation arising from the autoregressive approach.
The first factor implies that forecasts that are less smoothed tend to exhibit higher RMSE, known as the double-penalty effect \cite{Hoffman1995,Ebert2013}. 
This interpretation is supported by the ability of the proposed model 
to better capture the temporal evolution of fine-scale structures (Figure~\ref{fig:VS_XiHe_India}). 
This highlights the importance of conducting evaluations that do not rely solely on point-wise metrics such as RMSE. 
For example, \citeA{Bonavita2024} revealed that data-driven weather prediction models showed superiority in terms of RMSE but lacked physical consistency.
Although the double-penalty effect may play a role,
error accumulation during autoregressive rollout may also contribute to the degradation of predictions, 
leading to increased RMSE. 
Disentangling these factors and conducting further investigations are crucial for future research.

The evaluation presented in this study was conducted in a simplified setting,
which included relying on the reanalysis dataset and focusing on comparison with a data-driven model.
Regarding the first simplified setting, 
the evaluation requires comprehensive data that are spatially and temporally consistent, such as ocean reanalysis products,
as this study focuses on evaluating the representation of spatial distributions including fine-scale structures.
In addition, the initial conditions were obtained from the reanalysis dataset,
allowing the evaluation to focus directly on the capability of the model,
rather than on errors stemming from the initial state.
Regarding the second simplified setting, the evaluation focuses on a comparison with \citeA{Wang2024}, 
and no comparisons were made with state-of-the-art operational numerical ocean prediction systems.
This comparison is performed to reveal the distinctive characteristics of the proposed architecture,
such as the integration of a multi-scale GNN, an autoregressive formulation, and atmospheric forcing.
To this end, a comparison with another data-driven model trained on the same dataset provides a clearer basis for highlighting 
the advantages and limitations of the proposed model 
and is therefore suitable for the objectives of this study.

Despite the simplified setting, 
this study highlights the potential of the multi-scale GNN-based ocean model
for 10-day global ocean forecasting at eddy-resolving resolution.
Compared with the prior eddy-resolving data-driven global ocean model \cite{Wang2024} as a baseline,
the proposed model exhibits better representation of a wide range of spatial scales 
as well as temporal evolution, which are crucial for eddy-resolving ocean models,
though RMSE degradation has been observed at longer lead times.
In addition to its performance, a key advantage of the multi-scale GNN architecture lies in its flexibility,
particularly its ability to be extended through mesh configuration design.
For example, unstructured meshes, which are commonly used in computational fluid dynamics and adopted in ocean models such as MPAS-Ocean \cite{Ringler2013}, 
can be incorporated into the architecture, similar to GNN-based weather prediction models that use stretched grids \cite{Nipen2024}.
Such an extension would further advance data-driven ocean modeling and presents scope for future investigation.

\newpage

\begin{table}[htbp]
      \caption{Definition, description, and data sources of variables}
      \centering
      \begin{tabular}{l p{7.5cm} p{3cm}}
      \hline
      \textbf{Variable} & \textbf{Description} & \textbf{Data source} \\
      \hline
      $X^t$ & Temperature, eastward current, northward current, salinity, sea surface height$^{a}$. 
            & GLORYS \\
      $A^t$ & 10m eastward wind, 10m northward wind, precipitation, 2m temperature, 2m dewpoint temperature, shortwave radiation flux, longwave radiation flux, latent heat flux, sensible heat flux, sea level pressure.
            & ERA5 Reanalysis (training), GFS Forecasts (evaluation) \\
      $S$   & Latitude, longitude, ocean depth.
            & GLORYS \\
      \hline
      \multicolumn{3}{p{13cm}}{$^{a}$All variables except sea surface height include 23 vertical layers at
      0.49, 2.65, 5.08, 7.93, 11.41, 15.81, 21.60, 29.44, 40.34, 55.76, 77.85, 92.32, 109.73, 130.67, 155.85, 186.13, 222.48, 266.04, 318.13, 380.21, 453.94, 541.09, and 643.57~m.}
      \end{tabular}
      \label{tab:variables}
\end{table}

\clearpage

\section*{Data Availability Statement}
The GLORYS reanalysis dataset is available at \url{https://doi.org/10.48670/moi-00021}.  
The model by \citeA{Wang2024}, version released on 2024.09.22, is available from their GitHub repository:  
\url{https://github.com/Ocean-Intelligent-Forecasting/XiHe-GlobalOceanForecasting}.

\section*{Software Availability Statement}
Software supporting this research was developed under restrictions that include a required NDA, and is not publicly accessible.
The software can be made available to other researchers upon establishing an NDA with the authors’ affiliated institution.


%
%


%
%
%
%
%

\clearpage

\bibliography{manual}

\end{document}